\def\year{2021}\relax
\documentclass[letterpaper]{article} 
\usepackage{aaai21}  
\usepackage{times}  
\usepackage{helvet} 
\usepackage{courier}  
\usepackage[hyphens]{url}  
\usepackage{graphicx} 
\usepackage{natbib}  
\usepackage{caption} 
\usepackage[lofdepth,lotdepth]{subfig}
\usepackage[page]{appendix}
\usepackage{multirow}
\usepackage{array}

\title{Commonsense Knowledge Mining from Term Definitions}
\author{
    Zhicheng Liang,
    Deborah L. McGuinness\\
}
\affiliations{
    Department of Computer Science, Rensselaer Polytechnic Institute\\
    Troy, NY, USA\\
    liangz4@rpi.edu, dlm@cs.rpi.edu

}

\begin{document}
\maketitle

\begin{abstract}
Commonsense knowledge has proven to be beneficial to a variety of application areas, including question answering and natural language understanding. Previous work explored collecting commonsense knowledge triples automatically from text to increase the coverage of current commonsense knowledge graphs. We investigate a few machine learning approaches to mining commonsense knowledge triples using dictionary term definitions as inputs and provide some initial evaluation of the results. We start from extracting candidate triples using part-of-speech tag patterns from text, and then compare the performance of three existing models for triple scoring. Our experiments show that term definitions contain some valid and novel commonsense knowledge triples for some semantic relations, and also indicate some challenges with using existing triple scoring models.\footnote{Our code and data are available at: \url{https://github.com/gychant/CSKMTermDefn}}
\end{abstract}

\section{Introduction}
A variety of natural language related tasks, e.g. question answering~\cite{kagnet-emnlp19} and dialog systems~\cite{AAAI1816573}, are able to achieve better performance by introducing commonsense knowledge. However, the knowledge collection process is difficult because commonsense knowledge is assumed to be widely known, thus rarely stated explicitly in natural language text.
A Commonsense Knowledge Graph (CSKG) is usually represented as a directed graph, where nodes represent concepts and edges denote some pre-defined relations between concepts. Most existing, large-scale CSKGs are built by expert annotation, e.g. Cyc~\cite{lenat1995cyc}, or by crowdsourcing, e.g. ConceptNet~\cite{speer2017conceptnet} and ATOMIC~\cite{sap2019atomic}. With respect to the broadness and diversity of commonsense knowledge, these CSKGs typically suffer from low coverage.

To increase coverage, there have been two lines of research to infer commonsense knowledge automatically. One is CSKG completion, which aims to learn a scoring model that distinguishes between triples expressing commonsense knowledge and those that do not~\cite{li2016commonsense,saito2018commonsense,malaviya2020commonsense}. A learned scoring model can be used to estimate the plausibility of candidate concept-relation-concept triples that are either constructed between existing concepts or extracted from the raw text of new sources. Triples scored above a certain threshold are considered to be valid and then used to augment a CSKG.

The other line is CSKG generation, which aims to generate a new node (or, concept), $t_2$, in arbitrary phrases, and connect it with an existing node, $t_1$, using a pre-defined relation, $R$, in order to construct a new triple $(t_1, R, t_2)$~\cite{saito2018commonsense,Bosselut2019COMETCT}. 
One weakness of these generative models is the low novelty rate of generated concepts. As reported in the ConceptNet experiments by~\citet{Bosselut2019COMETCT}, only 3.75\% of generated triples contain novel object nodes given the subject and the relation of a triple, which limits the ability to augment a CSKG on a large scale. In contrast, the CSKG completion approach is more promising because it allows the introduction of more novel/unseen concepts to the existing graph, by mining diverse candidate triples from external text resources.

Previous work on commonsense knowledge mining exists.~\citet{blanco2011commonsense} extracted commonsense knowledge using concept properties and metarules.~\citet{li2016commonsense} extracted knowledge triples from Wikipedia and used their trained model to score the triples.~\citet{jastrzkebski2018commonsense} further evaluated the novelty of these extracted triples and introduced an automated novelty metric that correlates with human judgement.~\citet{davison-etal-2019-commonsense} leveraged a pre-trained language model to mine commonsense knowledge.~\citet{zhang2020TransOMCS} mined commonsense knowledge from linguistic patterns of raw text. However, as a potential source of commonsense knowledge, dictionary term definitions have yet to be explored.

Dictionary term definitions are compiled to provide precise descriptions of the properties of terms, concepts, or entities in our daily life. Based on the assumption that \textit{concepts have properties which imply commonsense}~\cite{blanco2011commonsense}, we further assume that some commonsense knowledge could be extracted or inferred from these definitions. For example, the term definition of ``bartender" is \textit{One who tends a bar or pub; a person preparing and serving drinks at a bar}, from which one could infer commonsense triples such as \emph{(bartender, IsA, person), (bartender, AtLocation, bar), (bartender, AtLocation, pub), (bartender, CapableOf, preparing and serving drinks)}. Among these triples, the second triple is already included in ConceptNet, and the others have semantically similar counterparts, e.g. \emph{(bartender, CapableOf, fill glass with drink), (bartender, CapableOf, mix drink)}, and \emph{(bartender, CapableOf, shake drink)}.

We aim to examine the performance of existing machine learning approaches for mining the described commonsense knowledge triples from such term definitions, i.e. examining their capability of distinguishing valid and invalid triples extracted from the raw text, and to understand the potential and feasibility for mining commonsense automatically from this particular kind of resource.

\section{Approach}
In this section, we introduce how we extract candidate triples of commonsense knowledge from term definitions, as well as the models we use to evaluate their plausibility scores, which measure the level of validity of a triple.

\subsection{Candidate Extraction}
We use term definitions from the English version of Wiktionary, a freely-available multilingual dictionary.\footnote{\url{https://en.wiktionary.org/wiki/Wiktionary:Main_Page}} Yet, our approach is agnostic to a particular definition resource. As in ConceptNet, the subject and object of a commonsense knowledge triple can be arbitrary phrases.
Instead of generating candidates from simple N-grams,~\citet{li2016commonsense} extracted candidates using frequent part-of-speech (POS) tag patterns of concept pairs for each pre-defined relation. This method encourages candidates to respect the target commonsense knowledge graph being extended. We employ a similar approach to extracting candidates from term definitions. First, we parse the nodes in ConceptNet using spaCy
to obtain their POS tags.\footnote{\url{https://spacy.io/}}
Next, we choose the top $k$ most frequent POS tag patterns for each relation and apply them to match text spans from term definition text. For instance, the phrase ``\textit{preparing and serving drinks}" can be extracted from the term definition of ``\textit{bartender}" using the POS tag pattern ``\textit{VERB},\textit{CCONJ},\textit{VERB},\textit{NOUN}" for the \textit{CapableOf} relation.
Finally, we construct knowledge triple candidates for each relation using a term as the subject and an extracted phrase from the definition of this term as the object.

\subsection{Triple Scoring}
We adopt three state-of-the-art triple scoring models for computing the plausibility of candidate triples. Each of them assigns a real-valued score to a given triple:
\begin{itemize}
    \item Bilinear AVG (i.e. average)~\cite{li2016commonsense} defines the plausibility score of a triple $(t_1, R, t_2)$ as $u_1^\top M_R u_2$ where $M_R \in \mathcal{R}^{r \times r}$ is the parameter matrix for relation $R$, $u_i$ is a nonlinear transformation of the term vector $v_i$, and $v_i$ is obtained by averaging the word embeddings of the original term, $t_i$. \citet{li2016commonsense} used this model to score triples extracted from Wikipedia, since it performs better when scoring novel triples.
    \item KG-BERT~\cite{yao2019kg} treats triples in knowledge graphs as textual sequences by taking the corresponding entity and relation descriptions, and learns the triple scoring function by fine-tuning a pre-trained language model.
    \item PMI model~\cite{davison-etal-2019-commonsense} represents triple plausibility using pointwise mutual information (PMI) between head entity, $\mathbf{h}$, and tail entity, $\mathbf{t}$. Specifically, it translates a relational triple into a masked sentence and estimates the PMI using a pre-trained language model, computed by $\textbf{PMI}(\mathbf{t}, \mathbf{h} | r) = \log p(\mathbf{t} | \mathbf{h}, r) - \log p(\mathbf{t} | r)$. The final plausibility score is obtained by averaging $\textbf{PMI}(\mathbf{t}, \mathbf{h} | r)$ and $\textbf{PMI}(\mathbf{h}, \mathbf{t} | r)$. This approach is unsupervised in the sense that the model does not need to be trained on a particular commonsense knowledge base to learn model weights. \citet{davison-etal-2019-commonsense} find that it outperforms supervised methods when mining commonsense knowledge from new sources.
\end{itemize}

\section{Experiment}
We provide details of candidate triple mining and our execution of the three models on the candidate triples, and then include some analysis of our results.

\subsection{Candidate Triple Mining}
ConceptNet 5~\cite{speer2017conceptnet}, the latest version of ConceptNet, is built from multiple sources, the Open Mind Common Sense (OMCS) project, Wiktionary, DBpedia, etc., thus introducing some domain-specific knowledge. To focus on commonsense knowledge only, we collect Wiktionary definitions for the terms that appeared in the English core of ConceptNet (total 162,363 terms).\footnote{We use the English triples with ConceptNet 4 as source.} We further filter out definitions related to morphology, i.e. the ones containing ``plural of", ``alternative form of", ``alternative spelling of", ``misspelling of", resulting in 13,850 term definitions. We choose 12 representative ConceptNet relations as extraction targets. By applying the top 15 frequent POS tag patterns for each relation, we extracted around 1.4 million candidate triples in total for these relations (see Table~\ref{tab:triple_stat} of the Appendix for detailed statistics).

\begin{figure*}[t]
\centering
\subfloat[Bilinear AVG]{
  \includegraphics[width=0.3\textwidth]{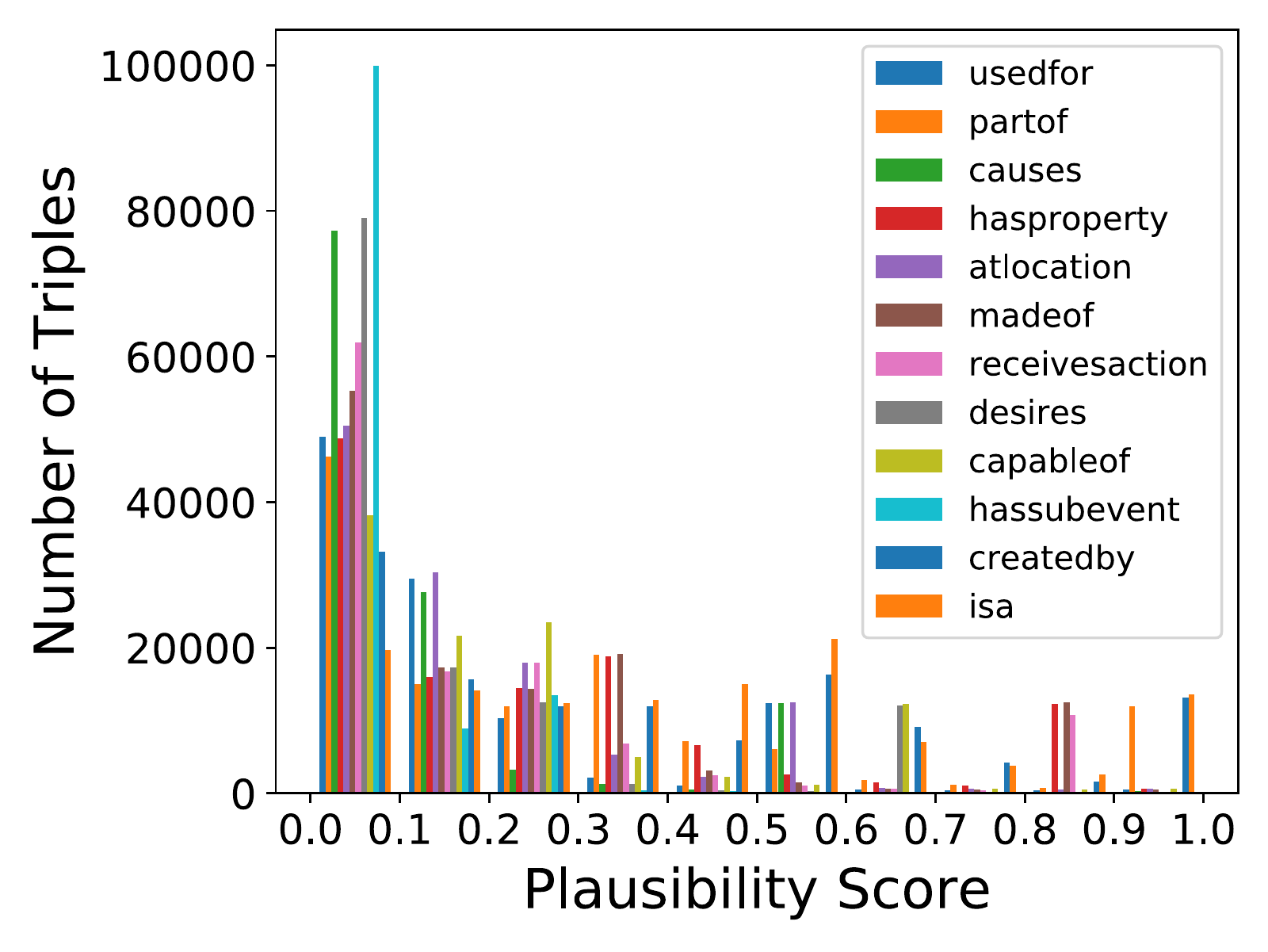}
  \label{fig:bilinear}}
\subfloat[KG-BERT]{
  \includegraphics[width=0.3\textwidth]{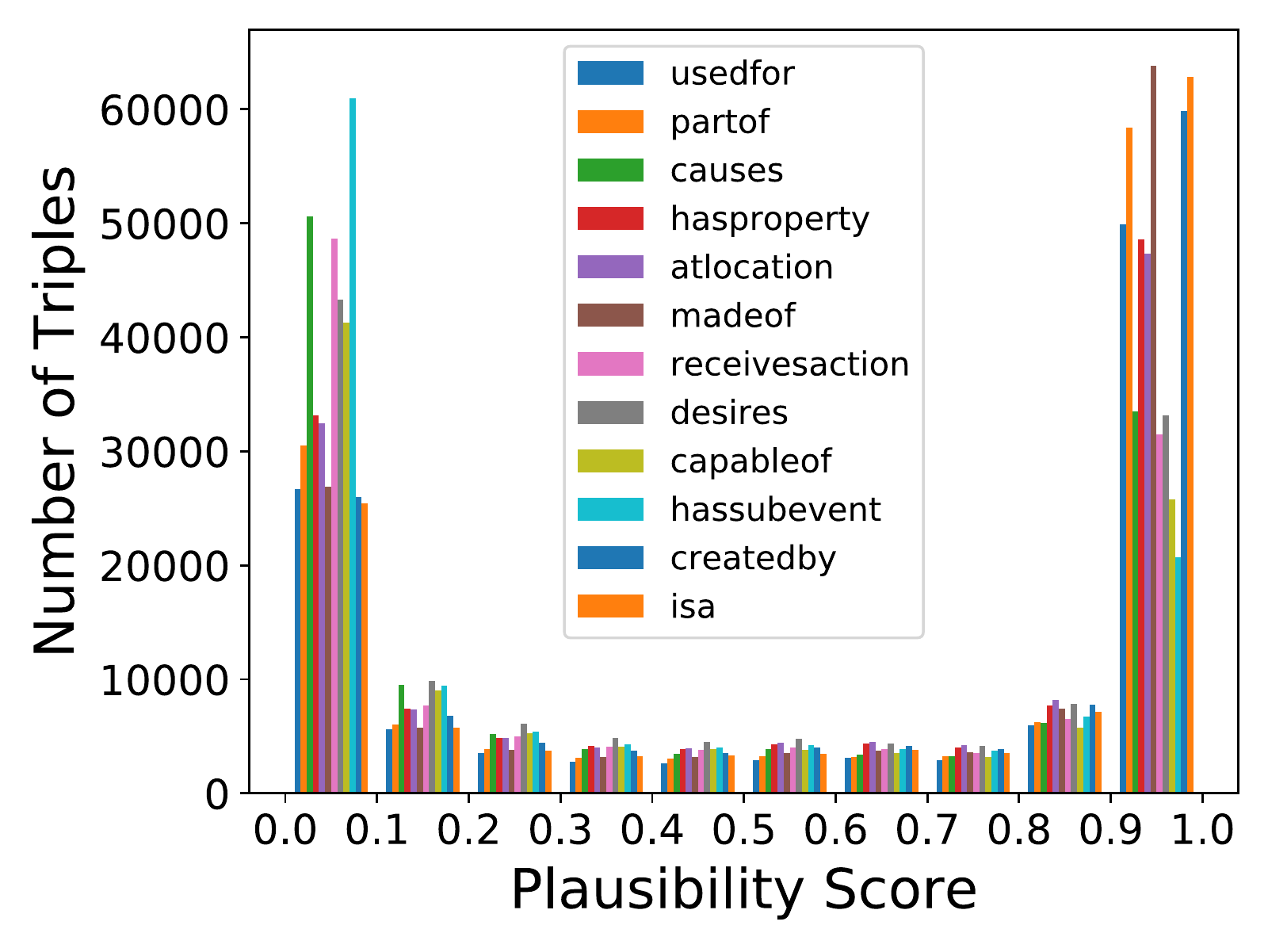}
  \label{fig:kgbert}}
\subfloat[PMI]{
  \includegraphics[width=0.3\textwidth]{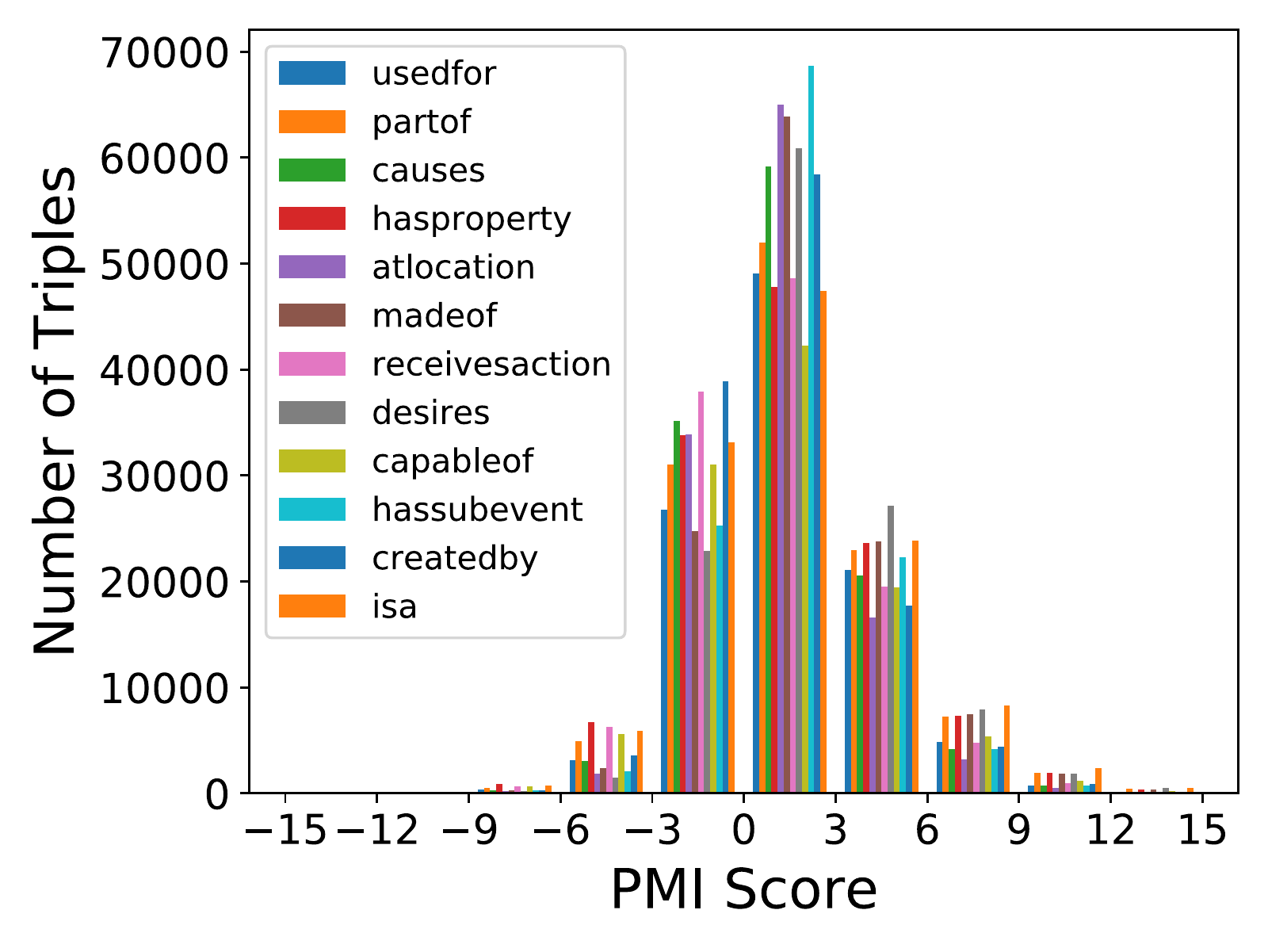}
  \label{fig:pmi}}
\caption{Score distribution of collected triples (best viewed in color).}
\label{fig:triple_scores}
\end{figure*}

\subsection{Model Settings}
For Bilinear AVG, we simply adopt the trained model released by~\citet{li2016commonsense}.\footnote{Available at~\url{https://ttic.uchicago.edu/~kgimpel/comsense_resources/ckbc-demo.tar.gz}} For KG-BERT,\footnote{Available at~\url{https://github.com/yao8839836/kg-bert}} we train the model using the full training set of 100 thousand triples that were used to train Bilinear AVG.\footnote{Available at~\url{https://ttic.uchicago.edu/~kgimpel/comsense\_resources/train100k.txt.gz}} Since this 100k training set contains only positive triples, KG-BERT generates negative triples by replacing the head entity or tail entity of a positive triple with a random entity in the training set. Note that if a corrupted triple with a replaced entity is already among the positive triples, it will not be treated as a negative triple. We train KG-BERT for 10 epochs and use the model that achieves the best accuracy on the development set released by~\citet{li2016commonsense}.\footnote{KG-BERT achieves 77.5\% accuracy on the dev set after 3 training epochs.} For PMI model, we directly run it on the candidate triples without training. Specifically, Bilinear AVG and KG-BERT assign each candidate triple a plausibility score in the range of [0, 1]. After running these models, we separately rank the candidate triples of each relation by their scores in descending order.

\subsection{Analysis}
We first analyze the distribution of scores assigned to the collected candidate triples by the three models we considered. The histogram plots with 10 bins are shown in Figure~\ref{fig:triple_scores}. We observe high variability in score distribution across these models, indicating that they do not have an evident consensus towards the plausibility of the candidate triples. It is also supported by the Kendall's tau coefficients~\cite{kendall1938new} computed between each pair of the models, whose absolute values are all less than 0.01. 

To evaluate the inferred knowledge, we adopt two metrics in the literature:
\textit{Validity} and \textit{Novelty}. Validity describes how many generated/extracted triples are plausible and is measured by human evaluation.~\citet{Bosselut2019COMETCT} conducted automatic evaluation using the pre-trained Bilinear AVG model developed by~\citet{li2016commonsense}. Secondly, novelty is measured by the percentage of generated triples that are unseen in the training set~\cite{li2016commonsense,saito2018commonsense}. We also evaluate it against the English subset of ConceptNet. Of the two, validity is more important, yet novelty is also a desirable property when augmenting a CSKG.

\noindent \textbf{1. Novelty} When we measure novelty of a triple, we treat concepts in the triple as strings. Preprocessing includes removing stop words in the strings, and treating them as a bag of words after lemmatization and stemming. This approach is stricter than exact match. Similarly,~\citet{Bosselut2019COMETCT} measured novelty using word token based minimum edit distance of generated phrase objects to the nearest one in the training set. We find that around 99$\%$ of the candidate triples of each relation are novel w.r.t either triples in the training set, or triples in the English core of ConceptNet 5. Notably, ConceptNet 5 already contains some triples whose concept pairs are automatically extracted from Wiktionary definitions and have been assigned a vague relation \textit{RelatedTo}, yet only for concept pairs both having a Wiktionary entry. Even with respect to the full English subset of ConceptNet 5, the novelty rates of candidate triples are above 99$\%$ for nearly all relations (except \textit{IsA} with 94.7$\%$). Notably, novelty may be affected by the presence of synonyms. To approximate novelty of triples based on semantic distance,~\citet{jastrzkebski2018commonsense} used the sum of Euclidean distances between the averaged word embeddings of the heads and the tails, respectively, of two triples. We tested this approach using our data but it seems not a good proxy for novelty. We leave further novelty analysis as future work.\\
\textbf{2. Validity} We sample 50 triples out of the high-scored triples of each relation and manually evaluate their validity as well as novelty using the aforementioned metric. The results are summarized in Table~\ref{tab:triple_quality}. Specifically, we sample triples from the ones scored at least 0.9 for Bilinear AVG and KG-BERT (see the \textit{Qual.} columns for the numbers of ``qualified" triples meeting this selection criterion), and sample from the top 1,000 high-scored triples for the PMI model since the scores are not in the range of [0, 1]. We report the proportion of valid triples in the samples (see the \textit{V.} columns), and also the proportion of triples being valid and novel (see the \textit{V.N.} columns). Table~\ref{tab:triple_example} of the Appendix lists some of the valid and novel triples from the samples we manually evaluated. The results show that the models do mine some valid and novel triples and the performance varies on different relations. Bilinear AVG achieves relatively high accuracy on relations e.g. \textit{HasProperty}, \textit{UsedFor}, \textit{AtLocation}, while getting only 6$\%$ accuracy on \textit{CausedBy}. KG-BERT performs better on some relations e.g. \textit{UsedFor}, \textit{CapableOf}, than other relations. PMI achieves better accuracy on the relation \textit{IsA} compared with the other two models. From the distribution of accuracy, we infer that term definitions of Wiktionary contain more commonsense knowledge for the relations with relatively higher accuracy, including \textit{AtLocation}, \textit{CapableOf}, \textit{HasProperty}, \textit{IsA}, \textit{MadeOf}, \textit{UsedFor}. This matches our observation that dictionary term definitions describe more on such basic properties of a concept.

\section{Discussion}
Our evaluations indicate the feasibility to mine commonsense triples using term definitions, for which we provide evidence using Wiktionary. We also find that the three models under evaluation have weaknesses in scoring triples. This is because the triples for manual evaluation are all sampled from the high-scored triples, yet for all relations, less than 50\% of the triples are valid. To some extent, the actual validity of candidate triples could be ruled out since invalid triples should have been assigned low scores (thus not in our samples for manual evaluation) if a model is sufficiently discriminatory. We do not conclude on which model is the best due to the large discrepancy of their score distributions.

\begin{table}[t]
{\small
\begin{tabular}{|@{\hspace{0.3\tabcolsep}}p{0.1\textwidth}|p{0.04\textwidth}|@{\hspace{0.6\tabcolsep}}p{0.02\textwidth}|@{\hspace{0.6\tabcolsep}}p{0.02\textwidth}|p{0.04\textwidth}|@{\hspace{0.6\tabcolsep}}p{0.02\textwidth}|@{\hspace{0.6\tabcolsep}}p{0.02\textwidth}|p{0.02\textwidth}|@{\hspace{0.6\tabcolsep}}p{0.02\textwidth}|} 
\hline
\multirow{2}{*}{\textbf{Relation}} & \multicolumn{3}{c|}{\textbf{Bilinear AVG}} & \multicolumn{3}{c|}{\textbf{KG-BERT}} & \multicolumn{2}{c|}{\textbf{PMI}} \\ 
\cline{2-9}
\multirow{2}{*}{} & Qual. & V. & V.N. & Qual. & V. & V.N. & V. & V.N. \\
\hline 
AtLocation & 632 & 0.44 & 0.30 & 47,315 & 0.04 & 0.04 & 0.10 & 0.10 \\ 
\hline
CapableOf & 585 & 0.30 & 0.28 & 25,771 & 0.26 & 0.26 & 0.12 & 0.12 \\ 
\hline
Causes & 293 & 0.18 & 0.16 & 33,511 & 0.02 & 0.02 & 0 & 0  \\ 
\hline
CreatedBy & 13,156 & 0.06 & 0.06 & 59,836 & 0.02 & 0.02 & 0.02 & 0.02 \\ 
\hline
Desires & 67 & 0.22 & 0.18 & 33,126 & 0 & 0 & 0.04 & 0.04  \\ 
\hline
HasProperty & 635 & 0.44 & 0.38 & 48,603 & 0.06 & 0.06 & 0.08 & 0.08 \\ 
\hline
HasSubevent & 173 & 0.10 & 0.08 & 20,719 & 0.06 & 0.06 & 0.02 & 0.02 \\ 
\hline
IsA & 13,537 & 0.36 & 0.28 & 62,819 & 0.24 & 0.16 & 0.46 & 0.34 \\ 
\hline
MadeOf & 487 & 0.26 & 0.24 & 63,781 & 0.14 & 0.12 & 0.10 & 0.08 \\ 
\hline
PartOf & 11,909 & 0.14 & 0.12 & 58,410 & 0.16 & 0.16 & 0.04 & 0.04 \\ 
\hline
ReceivesAction & 238 & 0.12 & 0.12 & 31,491 & 0.20 & 0.20 & 0.06 & 0.06 \\ 
\hline
UsedFor & 491 & 0.50 & 0.32 & 49,913 & 0.36 & 0.34 & 0.18 & 0.18  \\ 
\hline
\end{tabular}
}
\caption{Manual evaluation results of high-scored triples.}
\label{tab:triple_quality}
\end{table}
Our manual evaluation also uncovers some interesting model behaviors. Bilinear AVG tends to assign very high scores ($>$ 0.99) to the \textit{UsedFor} relation given concept pairs that actually have the \textit{IsA} relation. KG-BERT tends to assign very high scores to the \textit{CreatedBy} relation given concept pairs that actually have the \textit{CapableOf} relation. Regarding the validity of the candidate triples we extracted from Wiktionary, we can get some sense from the results of manual evaluation. For KG-BERT that have a large amount of triples scored above 0.9, as shown in Figure~\ref{fig:kgbert} and Table~\ref{tab:triple_quality}, we could roughly estimate the number of valid triples for each relation using the accuracy numbers in Table~\ref{tab:triple_quality}, e.g. around $50,000 \times 0.34 = 17,000$ plausible \textit{UsedFor} triples.

Since our analysis shows relatively wide variability in the studied scoring methods, reliance on these methods may need a deeper evaluation of their individual strengths and weaknesses. Further, validity and novelty are two useful metrics, however, additional research is needed to really consider when mined valid and novel content is worth adding. For instance, we obtained a valid triple~\emph{(camper, AtLocation, tent)} from the definition of~\emph{camper} - ``A person who camps, especially in a tent etc." - while ConceptNet already has a similar one \emph{(camper, CapableOf, sleep in a tent)}. To determine whether the former one is worth adding is challenging. One practical approach is to use the performance gain of downstream applications, like question answering, as the criterion for decision making.

\section{Conclusion and Future Work}
We presented a study on the feasibility of mining commonsense knowledge triples from Wiktionary term definitions. We examined three models on the performance of scoring newly extracted triples. We showed that they do mine some valid and novel triples and the performance varies on different semantic relations, and also observed some model weaknesses, e.g. low accuracy for high-scored triples and high variability between these models. Our findings suggest careful model pre-evaluation for use in practice. We plan to improve scoring models and candidate extraction techniques, and study the impact of new triples on downstream tasks.\\
\noindent\textbf{Acknowledgments} This work is funded through the DARPA MCS program award number N660011924033 to RPI under USC-ISI West.

\fontsize{9.0pt}{10.0pt} \selectfont
\bibliography{reference.bib}

\begin{thebibliography}{15}
\providecommand{\natexlab}[1]{#1}
\providecommand{\url}[1]{\texttt{#1}}
\providecommand{\urlprefix}{URL }
\expandafter\ifx\csname urlstyle\endcsname\relax
  \providecommand{\doi}[1]{doi:\discretionary{}{}{}#1}\else
  \providecommand{\doi}{doi:\discretionary{}{}{}\begingroup
  \urlstyle{rm}\Url}\fi

\bibitem[{Blanco, Cankaya, and Moldovan(2011)}]{blanco2011commonsense}
Blanco, E.; Cankaya, H.; and Moldovan, D. 2011.
\newblock Commonsense knowledge extraction using concepts properties.
\newblock In \emph{Twenty-Fourth International FLAIRS Conference}. Citeseer.

\bibitem[{Bosselut et~al.(2019)Bosselut, Rashkin, Sap, Malaviya, Çelikyilmaz,
  and Choi}]{Bosselut2019COMETCT}
Bosselut, A.; Rashkin, H.; Sap, M.; Malaviya, C.; Çelikyilmaz, A.; and Choi,
  Y. 2019.
\newblock COMET: Commonsense Transformers for Automatic Knowledge Graph
  Construction.
\newblock In \emph{ACL}.

\bibitem[{Davison, Feldman, and Rush(2019)}]{davison-etal-2019-commonsense}
Davison, J.; Feldman, J.; and Rush, A. 2019.
\newblock Commonsense Knowledge Mining from Pretrained Models.
\newblock In \emph{EMNLP-IJCNLP}.

\bibitem[{Jastrzebski et~al.(2018)Jastrzebski, Bahdanau, Hosseini, Noukhovitch,
  Bengio, and Cheung}]{jastrzkebski2018commonsense}
Jastrzebski, S.; Bahdanau, D.; Hosseini, S.; Noukhovitch, M.; Bengio, Y.; and
  Cheung, J. C.~K. 2018.
\newblock Commonsense mining as knowledge base completion? A study on the
  impact of novelty.
\newblock In \emph{Proc. of the Workshop on Generalization in the Age of Deep
  Learning}, 8--16.

\bibitem[{Kendall(1938)}]{kendall1938new}
Kendall, M.~G. 1938.
\newblock A new measure of rank correlation.
\newblock \emph{Biometrika} 30(1/2): 81--93.

\bibitem[{Lenat(1995)}]{lenat1995cyc}
Lenat, D.~B. 1995.
\newblock CYC: A large-scale investment in knowledge infrastructure.
\newblock \emph{Communications of the ACM} 38(11): 33--38.

\bibitem[{Li et~al.(2016)Li, Taheri, Tu, and Gimpel}]{li2016commonsense}
Li, X.; Taheri, A.; Tu, L.; and Gimpel, K. 2016.
\newblock Commonsense knowledge base completion.
\newblock In \emph{ACL}, 1445--1455.

\bibitem[{Lin et~al.(2019)Lin, Chen, Chen, and Ren}]{kagnet-emnlp19}
Lin, B.~Y.; Chen, X.; Chen, J.; and Ren, X. 2019.
\newblock KagNet: Knowledge-Aware Graph Networks for Commonsense Reasoning.
\newblock In \emph{EMNLP-IJCNLP}.

\bibitem[{Malaviya et~al.(2020)Malaviya, Bhagavatula, Bosselut, and
  Choi}]{malaviya2020commonsense}
Malaviya, C.; Bhagavatula, C.; Bosselut, A.; and Choi, Y. 2020.
\newblock Commonsense Knowledge Base Completion with Structural and Semantic
  Context.
\newblock \emph{AAAI} .

\bibitem[{Saito et~al.(2018)Saito, Nishida, Asano, and
  Tomita}]{saito2018commonsense}
Saito, I.; Nishida, K.; Asano, H.; and Tomita, J. 2018.
\newblock Commonsense knowledge base completion and generation.
\newblock In \emph{CoNLL}.

\bibitem[{Sap et~al.(2019)Sap, Le~Bras, Allaway, Bhagavatula, Lourie, Rashkin,
  Roof, Smith, and Choi}]{sap2019atomic}
Sap, M.; Le~Bras, R.; Allaway, E.; Bhagavatula, C.; Lourie, N.; Rashkin, H.;
  Roof, B.; Smith, N.~A.; and Choi, Y. 2019.
\newblock ATOMIC: An Atlas of Machine Commonsense for If-Then Reasoning.
\newblock In \emph{AAAI}, volume~33, 3027--3035.

\bibitem[{Speer, Chin, and Havasi(2017)}]{speer2017conceptnet}
Speer, R.; Chin, J.; and Havasi, C. 2017.
\newblock ConceptNet 5.5: an open multilingual graph of general knowledge.
\newblock In \emph{AAAI}, 4444--4451.

\bibitem[{Yao, Mao, and Luo(2019)}]{yao2019kg}
Yao, L.; Mao, C.; and Luo, Y. 2019.
\newblock KG-BERT: BERT for knowledge graph completion.
\newblock \emph{arXiv preprint arXiv:1909.03193} .

\bibitem[{Young et~al.(2018)Young, Cambria, Chaturvedi, Zhou, Biswas, and
  Huang}]{AAAI1816573}
Young, T.; Cambria, E.; Chaturvedi, I.; Zhou, H.; Biswas, S.; and Huang, M.
  2018.
\newblock Augmenting End-to-End Dialogue Systems With Commonsense Knowledge.
\newblock In \emph{AAAI}.

\bibitem[{Zhang et~al.(2020)Zhang, Khashabi, Song, and
  Roth}]{zhang2020TransOMCS}
Zhang, H.; Khashabi, D.; Song, Y.; and Roth, D. 2020.
\newblock TransOMCS: From Linguistic Graphs to Commonsense Knowledge.
\newblock In \emph{IJCAI}.

\end{thebibliography}

\normalsize
\onecolumn
\clearpage
\newpage
\renewcommand*\appendixpagename{Appendix}
\appendixpage
In Table~\ref{tab:triple_example}, we show some valid and novel knowledge triples from the samples we evaluated manually. We group them by the semantic relation and the model used to score them. In Table~\ref{tab:triple_stat}, we report the statistics of candidate triples.
\begin{table}[h]
\centering
\begin{tabular}{|c|p{0.25\textwidth}|p{0.25\textwidth}|p{0.25\textwidth}|} 
\hline
\multirow{2}{*}{\textbf{Relation}} & \multicolumn{3}{c|}{\textbf{Some Valid and Novel Examples from Human Evaluation}} \\
\cline{2-4}
\multirow{2}{*}{} & Bilinear AVG & KG-BERT & PMI \\
\hline 
AtLocation & \small{(camper, tent), (waiter, restaurant), (database, computer), (stove, room), (locker, store)} & \small{(paddler, canoe), (circle, figure)} & \small{(glioma, brain), (coalminer, coal), (collagen, extracellular), (thyroiditis, thyroid), (pneumothorax, chest)} \\ 
\hline
CapableOf & \small{(nose, smell), (owl, prey), (campfire, heat), (tablecloth, cover), (labor, work)} & \small{(salmonella, poisoning), (negotiation, achieving agreement), (shield, defense), (auto mechanic, repairing), (yeast, brew)} & \small{(pursuer, pursues), (massage therapist, massage therapy), (showroom, display), (passenger train, rail transport), (droplet, drop)} \\
\hline
Causes & \small{(damage, harm), (invest, development), (entertainment, enjoyment), (ponder, thought), (howl, sound)} & \small{(multiple sclerosis, depression)} & \small{N/A} \\ 
\hline
CreatedBy & \small{(corn earworm, helicoverpa zea), (kibbutz, economical sharing), (yale, heraldry)} & \small{(marine, military)} & \small{(noise pollution, excess noise)} \\ 
\hline
Desires & \small{(scientist, answer), (predator, prey), (graduate, degree), (workaholic, work), (judge, justice)} & \small{N/A} & \small{(sexist, practises sexism), (contestant, game show)} \\ 
\hline
HasProperty & \small{(fingerprint, unique), (reptile, cold-blooded), (chili, pungent), (beauty, attractive), (deck, flat)} & \small{(comet, celestial), (ratafia, bitter), (mashed potato, pulpy)} & \small{(forest, uncultivated), (copyright infringement, unauthorized use), (stockholder, owns stock), (beryllium, alkaline)}\\ 
\hline
HasSubevent & \small{(golf, hit), (bribe, exchange), (asthma, breath), (yawn, breath)} & \small{(meningitis, stiffness), (archery, shooting), (rebirth, birth)} & \small{(cheating, imposture)}\\ 
\hline
IsA & \small{(sailor suit, clothing style), (oil platform, large structure), (somnoplasty, medical treatment), (scanner, device), (immune response, integrated response)} & \small{(puma, lion), (mayor, leader), (royal, person), (bacteriostat, chemical)} & \small{(trombone, instrument), (fire truck, vehicle), (carnivore, animal), (cosmetic surgery, medical treatment), (katakana, Japanese syllabary)} \\ 
\hline
MadeOf & \small{(pickle, salt), (sushi, rice), (candle, wax), (cymbal, bronze), (vodka, grain)} & \small{(press release, statement), (daytime, time), (confetti, metal), (gas giant, methane), (casino, room)} & \small{(chocolate milk, milk), (coronary thrombosis, blood), (glassware, glass), (chopped liver, chicken liver)} \\ 
\hline
PartOf & \small{(expertise, knowledge), (barney, pejorative slang), (massage therapist, massage therapy)} & \small{(snooker table, snooker), (seabed, sea), (exotica, american music), (surcharge, price), (free market, market)} & \small{(chemistry, natural science), (barrier reef, adjacent coast)} \\ 
\hline
ReceivesAction & \small{(quantity, count), (speech, speak), (asset, value), (cleaner, clean), (plunger, remove)} & \small{(sausage, made), (experience, produced), (fishing rod, used), (ledger, record), (harpsichord, tuned)} & \small{(saddlery, saddler), (rowboat, rowing), (space station, habitation)} \\ 
\hline
UsedFor & \small{(message, communication), (mouthwash, clean), (ribbon, decoration), (tablecloth, protect), (article, report)} & \small{(hypothesis, observation), (immune system, response), (grader, maintenance), (nature reserve, conserve wildlife), (harpsichord, baroque music)} & \small{(meteorology, forecasting), (kidney, producing urine), (flush toilet, flush urine), (answer, question), (machine tool, machining)}\\ 
\hline
\end{tabular}
\caption{Examples of valid and novel triples.}
\label{tab:triple_example}
\end{table}

\begin{table}[h!]
\centering
\begin{tabular}{|c|c|c|c|c|c|c|}
\hline
Relation & AtLocation & CapableOf & Causes  & CreatedBy & Desires & HasProperty \\
\hline
\# of Candidates & 121,184 & 105,488 & 122,879 & 124,064 & 122,748 & 122,382 \\
\hline
\hline
Relation & HasSubevent & IsA & MadeOf & PartOf & ReceivesAction & UsedFor \\
\hline
\# of Candidates & 123,324 & 122,182 & 124,661 & 120,915 & 118,604 & 105,870 \\
\hline
\end{tabular}
\caption{Statistics of candidate triples.}
\label{tab:triple_stat}
\end{table}

\end{document}